\documentclass[11pt,a4paper]{article}
\usepackage{authblk}
\usepackage[hyperref]{acl2019}
\usepackage{times}
\usepackage{latexsym}
\usepackage{url}

\aclfinalcopy 
\def\aclpaperid{2206} 

\usepackage{url}
\usepackage{booktabs} 
\usepackage{graphicx}
\usepackage{epstopdf}
\usepackage{subfigure}
\usepackage{verbatim}
\usepackage{amssymb}
\usepackage{amsmath}
\usepackage{bm}
\usepackage{array}
\usepackage{multirow}
\usepackage{subfigure}

\newcommand{\nop}[1]{}

\newcolumntype{L}[1]{>{\raggedright\let\newline\\\arraybackslash\hspace{0pt}}m{#1}}
\newcolumntype{C}[1]{>{\centering\let\newline\\\arraybackslash\hspace{0pt}}m{#1}}
\newcolumntype{R}[1]{>{\raggedleft\let\newline\\\arraybackslash\hspace{0pt}}m{#1}}

\newenvironment{formalfont}{\fontfamily{pcr}\selectfont\small}{\par}
\DeclareTextFontCommand{\textformal}{\formalfont}

\title{Global Textual Relation Embedding for Relational Understanding}

\author[1]{\textbf{Zhiyu Chen}}
\author[1]{\textbf{Hanwen Zha}}
\author[1]{\textbf{Honglei Liu}}
\author[1]{\textbf{Wenhu Chen}}
\author[1]{\textbf{Xifeng Yan}}
\author[2]{\textbf{Yu Su}}
\affil[1]{University of California, Santa Barbara, CA, USA}
\affil[2]{The Ohio State University, OH, USA \authorcr \{zhiyuchen, hwzha, honglei, wenhuchen, xyan\}@cs.ucsb.edu, su.809@osu.edu}

\date{}

\begin{document}
\maketitle

\begin{abstract}
Pre-trained embeddings such as word embeddings and sentence embeddings are fundamental tools facilitating a wide range of downstream NLP tasks. In this work, we investigate how to learn a general-purpose embedding of textual relations, defined as the shortest dependency path between entities. 
Textual relation embedding provides a level of knowledge between word/phrase level and sentence level, and we show that it can facilitate downstream tasks requiring relational understanding of the text. To learn such an embedding, we create the largest distant supervision dataset by linking the entire English ClueWeb09 corpus to Freebase. We use global co-occurrence statistics between textual and knowledge base relations as the supervision signal to train the embedding. Evaluation on two relational understanding tasks demonstrates the usefulness of the learned textual relation embedding. The data and code can be found at \href{https://github.com/czyssrs/GloREPlus}{https://github.com/czyssrs/GloREPlus}

\end{abstract} 
\section{Introduction}
\label{sec:introduction}

Pre-trained embeddings such as word embeddings \cite{mikolov2013distributed,pennington2014glove,peters2018deep,devlin2018bert} and sentence embeddings \cite{le2014distributed,kiros2015skip} have become fundamental NLP tools. 
Learned with large-scale (e.g., up to 800 billion tokens \cite{pennington2014glove}) open-domain corpora, such embeddings serve as a good prior for a wide range of downstream tasks by endowing task-specific models with general lexical, syntactic, and semantic knowledge.

Inspecting the spectrum of granularity, a representation between lexical (and phrasal) level and sentence level is missing. 
Many tasks require relational understanding of the entities mentioned in the text, e.g., relation extraction and knowledge base completion. 
Textual relation \cite{bunescu2005shortest}, defined as the shortest path between two entities in the dependency parse tree of a sentence, has been widely shown to be the main bearer of relational information in text and proved effective in relation extraction tasks~\cite{xu2015classifying,su2017global}. 
If we can learn a \textit{general-purpose embedding for textual relations}, it may facilitate many downstream relational understanding tasks by providing general relational knowledge.

\begin{figure*}[htbp]
\begin{minipage}[t]{0.53\textwidth}
\centering
\vspace{0pt}
\includegraphics[width=3in]{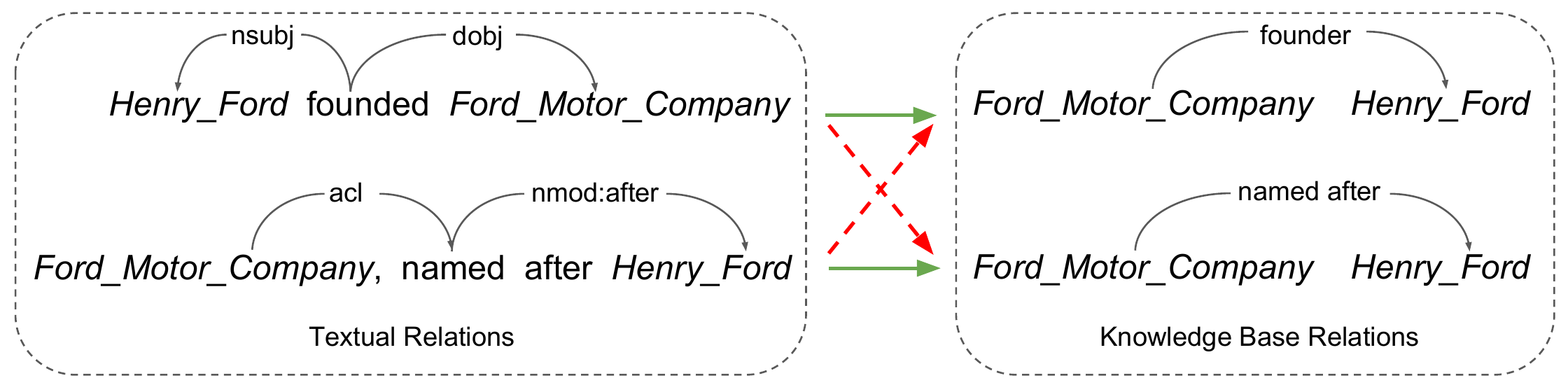}
\end{minipage}\quad
\begin{minipage}[t]{0.44\textwidth}
\centering
\vspace{1pt}
\resizebox{\columnwidth}{!}{%
\begin{tabular}{lcc}
\toprule
& $\xleftarrow{dobj} founded \xrightarrow{nsubj}$ & $\xrightarrow{acl} named \xrightarrow{nmod:after}$ \\
\midrule
founder & 2468.0 & 24.0 \\
named\_after & 305.0 & 347.0\\
... & ... & ... \\
\bottomrule
\end{tabular}
}
\end{minipage}
\caption{\textit{Left:} The wrong labeling problem of distant supervision. The Ford Motor Company is both founded by and named after Henry Ford. The KB relation \textit{founder} and \textit{named\_after} are thus both mapped to all of the sentences containing the entity pair, resulting in many wrong labels (red dashed arrows). \textit{Right:} Global co-occurrence statistics from our distant supervision dataset, which clearly distinguishes the two textual relations.} 
\label{figure:global_statics}
\end{figure*}

Similar to language modeling for learning general-purpose word embeddings, distant supervision \cite{mintz2009distant} is a promising way to acquire supervision, at no cost, for training general-purpose embedding of textual relations. Recently Su et al. \shortcite{su2017global} propose to leverage global co-occurrence statistics of textual and KB relations to learn embeddings of textual relations, and show that it can effectively combat the wrong labeling problem of distant supervision (see Figure~\ref{figure:global_statics} for example). While their method, named GloRE, achieves the state-of-the-art performance on the popular New York Times (NYT) dataset~\cite{riedel2010modeling}, the scope of their study is limited to relation extraction with small-scale in-domain training data.

In this work, we take the GloRE approach further and apply it to large-scale, domain-independent data labeled with distant supervision, with the goal of learning general-purpose textual relation embeddings. Specifically, we create the largest ever distant supervision dataset by linking the entire English ClueWeb09 corpus (half a billion of web documents) to the latest version of Freebase~\cite{bollacker2008freebase}, which contains 45 million entities and 3 billion relational facts. After filtering, we get a dataset with over 5 million unique textual relations and around 9 million co-occurring textual and KB relation pairs. We then train textual relation embedding on the collected dataset in a way similar to \cite{su2017global}, but using Transformer \cite{vaswani2017attention} instead of vanilla RNN as the encoder for better training efficiency.


To demonstrate the usefulness of the learned textual relation embedding, we experiment on two relational understanding tasks, relation extraction and knowledge base completion. For relation extraction, we use the embedding to augment PCNN+ATT~\cite{lin2016neural}\nop{ and improve the best $F_1$-score from 42.7\% to 44.6\%,} and improve the precision for top 1000 predictions from 83.9\% to 89.8\%. For knowledge base completion, we replace the neural network in \cite{toutanova2015representing} with our pre-trained embedding followed by a simple projection layer, and gain improvements on both MRR and HITS@10 measures. 
Our major contributions are summarized as following:
\begin{itemize}
\item We propose the novel task of learning general-purpose embedding of textual relations, which has the potential to facilitate a wide range of relational understanding tasks.
\item To learn such an embedding, we create the largest distant supervision dataset by linking the entire English ClueWeb09 corpus to Freebase. The dataset is publicly available\footnote{\href{https://github.com/czyssrs/GloREPlus}{https://github.com/czyssrs/GloREPlus}}. 
\item Based on the global co-occurrence statistics of textual and KB relations, we learn a textual relation embedding on the collected dataset and demonstrate its usefulness on relational understanding tasks. 
\end{itemize}
\section{Related Work}
\label{sec:related}
Distant supervision methods \cite{mintz2009distant} for relation extraction have been studied by a number of works \cite{riedel2010modeling,hoffmann2011knowledge,surdeanu2012multi,zeng2015distant,lin2016neural,ji2017distant,wu2017adversarial}.
\cite{su2017global} use global co-occurrence statistics of textual and KB relations to effectively combat the wrong labeling problem. But the global statistics in their work is limited to NYT dataset, capturing domain-specific distributions.


Another line of research that relates to ours is the universal schema~\cite{riedel2013relation} for relation extraction, KB completion, as well as its extensions~\cite{toutanova2015representing,verga2016multilingual}. 
Wrong labeling problem still exists since their embedding is learned based on individual relation facts. In contrast, we use the global co-occurrence statistics as explicit supervision signal. 


\section{Textual Relation Embedding}
\label{sec:model}

In this section, we describe how to collect large-scale data via distant supervision (\S\ref{sec:global_statistics}) and train the textual relation embedding  (\S\ref{sec:embedding_model}).

\subsection{Global Co-Occurrence Statistics from Distant Supervision}
\label{sec:global_statistics}
To construct a large-scale distant supervision dataset, we first get the English ClueWeb09 corpus~\cite{callan2009clueweb09}, which contains 500 million web documents. We employ the FACC1 dataset~\cite{gabrilovich2013facc1} to map ClueWeb09 to Freebase. We identify over 5 billion entity mentions in ClueWeb09 and link them to Freebase entities. From the linked documents, we extract 155 million sentences containing at least two entity mentions. We then use the Stanford
Parser~\cite{chen2014fast} with universal dependencies to extract textual relations (shortest dependency paths) between each pair of entity mentions\footnote{To be more precise, only shortest dependency paths without any other entity on the path are extracted.}, leading to 788 million relational triples (subject, textual relation, object), of which 451 million are unique.

Following \cite{su2017global}, we then collect the global co-occurrence statistics of textual and KB relations. More specifically, for a relational triple $(e_1, t, e_2)$ with textual relation $t$, if $(e_1, r, e_2)$ with KB relation $r$ exists in the KB, then we count it as a co-occurrence of $t$ and $r$. We count the total number of co-occurrences of each pair of textual and KB relation across the entire corpus. We then normalize the global co-occurrence statistics such that each textual relation has a valid probability distribution over all the KB relations, which presumably captures the semantics of the textual relation. In the end, a bipartite relation graph is constructed, with one node set being the textual relations, the other node set being the KB relations, and the weighted edges representing the normalized global co-occurrence statistics. 

\textbf{Filtering.} When aligning the text corpus with the KB, we apply a number of filters to ensure data quality and training efficiency: (1) We only use the KB relations in Freebase Commons, 70 domains that are manually verified to be of release quality. (2) Only textual relations with the number of tokens (including both lexical tokens and dependency relations) less than or equal to 10 are kept. (3) Only non-symmetric textual relations are kept, because symmetric ones are typically from conjunctions like "and" or "or", which are less of interest. (4) Only textual relations with at least two occurrences are kept. After filtering, we end up with a relation graph with \emph{5,559,176} unique textual relations, \emph{1,925} knowledge base (KB) relations, and \emph{8,825,731} edges with non-zero weight. It is worth noting that these filters are very conservative, and we can easily increase the scale of data by relaxing some of the filters.  

\nop{We extracted 788036618 sentences from ClueWeb09 and linked them to 2667688 unique entities. Each sentence contains at least two entities. }
\subsection{Embedding Training}
\label{sec:embedding_model}
Considering both effectiveness and efficiency, we employ the Transformer encoder \cite{vaswani2017attention} to learn the textual relation embedding. It has been shown to excel at learning general-purpose representations \cite{devlin2018bert}.

The embedded textual relation token sequence is fed as input. For example, for the textual relation $\xleftarrow{dobj} founded \xrightarrow{nsubj}$, the input is the embedded sequence of $\{<-dobj>, founded, <nsubj>\}$. We project the output of the encoder to a vector $z$ as the result embedding.
Given a textual relation $t_i$ and its embedding $z_i$, denote $\{r_1, r_2, ..., r_n\}$ as all KB relations, and $\tilde{p}(r_j|t_i)$ as the global co-occurrence distribution, the weight of the edge between textual relation $t_i$ and KB relation $r_j$ in the relation graph.
The training objective is to minimize the cross-entropy loss:
\begin{equation}
    \setlength{\abovedisplayskip}{4pt}
    \setlength{\belowdisplayskip}{-7pt}
L = -\sum_{i,j}\tilde{p}(r_j|t_i)log(p(r_j|t_i)),
\end{equation}
Where 
\begin{equation}
    \setlength{\abovedisplayskip}{2pt}
    \setlength{\belowdisplayskip}{5pt}
p(r_j|t_i) = (softmax(Wz_i + b))_j.
\end{equation}
$W$ and $b$ are trainable parameters. 

We use the filtered relation graph in \S\ref{sec:global_statistics} as our training data. To guarantee that the model generalizes to unseen textual relations, we take $5\%$ of the training data as validation set. Word embeddings are initialized with the GloVe \cite{pennington2014glove} vectors\footnotemark. Dependency relation embeddings are initialized randomly. 
\footnotetext{\href{https://nlp.stanford.edu/projects/glove/}{https://nlp.stanford.edu/projects/glove/}}
For the Transformer model, we use 6 layers and 6 attention heads for each layer. We use the Adam optimizer~\cite{kingma2014adam} with parameter settings suggested by the original Transformer paper~\cite{vaswani2017attention}. We train a maximum number of 200 epochs and take the checkpoint with minimum validation loss for the result. 

We also compare with using vanilla RNN in GloRE~\cite{su2017global}. Denote the embedding trained with Tranformer as \textbf{GloRE++}, standing for both new data and different model, and with RNN as \textbf{GloRE+}, standing for new data. We observe that, in the early stage of training, the validation loss of RNN decreases faster than Transformer. However, it starts to overfit soon.

\section{Experiments}
In this section, we evaluate the usefulness of the learned textual relation embedding on two popular relational understanding tasks, relation extraction and knowledge base completion. \emph{We do not fine-tune the embedding}, and only use in-domain data to train a single feedforward layer to project the embedding to the target relations of the domain. We compare this with models that are specifically designed for those tasks and trained using in-domain data. If we can achieve comparable or better results, it demonstrates that the general-purpose embedding captures useful information for downstream tasks.



\subsection{Relation Extraction}
\label{sec:relation_extraction}

\begin{table}[t]
\begin{center}
\resizebox{.48\textwidth}{!}{%
\begin{tabular}{lcccccc}
\toprule
Precision@N & 100 & 300 & 500 & 700 & 900 & 1000\\
\midrule
PCNN+ATT & 97.0 & 93.7 &	92.8 &	89.1 &	85.2 &	83.9\\
PCNN+ATT+GloRE & 97.0 & 97.3 & 94.6 & 93.3 & 90.1 & 89.3\\
\midrule
PCNN+ATT+GloRE+ & \textbf{98.0} & \textbf{98.7} & \textbf{96.6} & 93.1 & 89.9 & 88.8\\
PCNN+ATT+GloRE++ & \textbf{98.0} & 97.3 & 96.0 & \textbf{93.6} & \textbf{91.0} & \textbf{89.8}\\
\bottomrule
\end{tabular}
}
\end{center}
\caption{Relation extraction manual evaluation results: Precision of top 1000 predictions.}
\label{table:re_manual}
\end{table}
\begin{table*}[ht]
\begin{center}
\resizebox{.70\textwidth}{!}{%
\begin{tabular}{lcccccc}
\toprule
\multirow{2}{*}{Model} &
\multicolumn{2}{c}{Overall} &
\multicolumn{2}{c}{With mentions} &
\multicolumn{2}{c}{Without mentions}\\
\cline{2-7}
  & MRR & HITS@10 & MRR & HITS@10 & MRR & HITS@10 \\
 \hline
\hline
\textsc{DistMult} (KB only) & 35.8 & 51.8 & 27.3 & 39.5 & 39 & 56.3 \\
\hline
Conv-\textsc{DistMult} & 36.5 & 52.5 & 28.5 & 41.4 & 39.4 & 56.5 \\
\hline
Emb-\textsc{DistMult} (GloRE+) & 36.4 & 52.6 & \textbf{28.8} & \textbf{41.8} & 39.3 & 56.7 \\
\hline
Emb-\textsc{DistMult} (GloRE++) & \textbf{36.6} & \textbf{53.0} & 28.0 & 40.8 & \textbf{39.8} & \textbf{57.1} \\
\hline
\hline
\textsc{E+DistMult} (KB only) & 37.8 & 53.5 & 29.5 & 43 & 40.9 & 57.3 \\
\hline
Conv-\textsc{E}+Conv-\textsc{DistMult} & 38.7 & \textbf{54.4} & \textbf{30.0} & \textbf{43.8} & 41.9 & 58.2 \\
\hline
Emb-\textsc{E}+Emb-\textsc{DistMult} (GloRE+) & 38.8 & 54.2 & \textbf{30.0} & 43.3 & 42.0 & 58.2 \\
\hline
Emb-\textsc{E}+Emb-\textsc{DistMult} (GloRE++) & \textbf{38.9} & \textbf{54.4} & \textbf{30.0} & 43.5 & \textbf{42.1} & \textbf{58.3} \\
\bottomrule
\end{tabular}
}
\end{center}
\caption{Results of KB completion on FB15k-237 dataset\footnotemark, measured by MRR and HITS@10 (Both scaled by 100).}
\label{table:kbc_result}
\end{table*}
\newcommand{\tabincell}[2]{\begin{tabular}{@{}#1@{}}#2\end{tabular}}  
\begin{table*}[!htbp]
\centering
\resizebox{\textwidth}{!}{%
\small
\begin{tabular}{m{6cm}m{7cm}m{3cm}m{6.5cm}}
\toprule
Subject and object & \multicolumn{3}{l}{Francis Clark Howell, Kansas City} \\
\hline
KB relation & \multicolumn{3}{l}{people.person.place\_of\_birth} \\
\hline
Textual relation in NYT train set & \multicolumn{3}{l}{$\xleftarrow{nsubjpass} born \xrightarrow{nmod:on} nov. \xrightarrow{nmod:in}$} \\
\hline
Corresponding sentence in NYT train set &  \multicolumn{3}{l}{...\textcolor{red}{Francis Clark Howell} was born on nov. 27, 1925, in \textcolor{red}{Kansas City}, ...} \\
\hline
\hline
\multirow{3}{*}{Top-5 nearest neighbors in ClueWeb train set} & Textual relation & Cosine similarity & A corresponding sentence in ClueWeb raw data \\ 

\cline{2-4} & $\xleftarrow{nsubjpass} born \xrightarrow{nmod:in} 1295 \xrightarrow{nmod:in}$ & 0.61 & \tabincell{l}{...According to the Lonely Planet Guide to Venice, \\ \textcolor{red}{St. Roch} was born in 1295 in \textcolor{red}{Montpellier}, France, \\ and at the age of 20 began wandering...} \\

\cline{2-4} & $\xleftarrow{nsubjpass} born \xrightarrow{nmod:in} 1222 \xrightarrow{nmod:in}$ & 0.61 & \tabincell{l}{...\textcolor{red}{Isabel BIGOD} was born in 1222 in \textcolor{red}{Thetford} \\ \textcolor{red}{Abbey}, Norfolk, England...} \\

\cline{2-4} & $\xleftarrow{nsubjpass} born \xrightarrow{dobj} Lannerback \xrightarrow{nmod:in}$ & 0.60 & \tabincell{l}{...\textcolor{red}{Yngwie (pronounced "ING-vay") Malmsteen} was \\ born Lars Johann Yngwie Lannerback in \\ \textcolor{red}{Stockholm}, Sweden, in 1963, ...}\\

\cline{2-4} & $\xleftarrow{nsubjpass} born \xrightarrow{nmod:in} Leigha \xrightarrow{appos} Muzaffargarh \xrightarrow{nmod:in}$ & 0.57 & \tabincell{l}{...Satya Paul - Indian Designer \textcolor{red}{Satya Paul} was born \\ in Leigha, Muzaffargarh in \textcolor{red}{Pakistan},  and came to \\ India during the partition times...}\\

\cline{2-4} & $\xleftarrow{nsubjpass} born \xrightarrow{nmod:on} raised \xrightarrow{nmod:in}$ & 0.55 & \tabincell{l}{...\textcolor{red}{Governor Gilmore} was born on October 6, 1949 \\ and raised in \textcolor{red}{Richmond}, Virginia...}\\

\bottomrule
\end{tabular}
}

\caption{Case study: Textual relation embedding model can well generalize to unseen textual relations via capturing common shared sub-structures. }
\label{table:case_2}
\end{table*}

We experiment on the popular New York Times (NYT) relation extraction dataset~\cite{riedel2010modeling}. 
Following GloRE \cite{su2017global}, 
we aim at augmenting existing relation extractors with the textual relation embeddings. 
We first average the textual relation embeddings of all contextual sentences of an entity pair, and project the average embedding to the target KB relations. We then construct an ensemble model by a weighted combination of predictions from the base model and the textual relation embedding. 

Same as \cite{su2017global}, we use PCNN+ATT \cite{lin2016neural} as our base model. GloRE++ improves its best $F_1$-score from 42.7\% to 45.2\%, slightly outperforming the previous state-of-the-art (GloRE, 44.7\%). 
As shown in previous work \cite{su2017global}, on NYT dataset, due to a significant amount of false negatives, the PR curve on the held-out set may not be an accurate measure of performance. Therefore, we mainly employ manual evaluation. We invite graduate students to check top 1000 predictions of each method. They are present with the entity pair, the prediction, and all the contextual sentences of the entity pair. Each prediction is examined by two students until reaching an agreement after discussion. Besides, the students are not aware of the source of the predictions. 
Table \ref{table:re_manual} shows the manual evaluation results. Both GloRE+ and GloRE++ get improvements over GloRE. GloRE++ obtains the best results for top 700, 900 and 1000 predictions.
\subsection{Knowledge Base Completion}
\label{sec:kbc}
We experiment on another relational understanding task, knowledge base (KB) completion, on the popular FB15k-237 dataset \cite{toutanova2015representing}. The goal is to predict missing relation facts based on a set of known entities, KB relations, and textual mentions. 
\cite{toutanova2015representing} use a convolutional neural network (CNN) to model textual relations. We replace their CNN with our pre-trained embedding followed by one simple feed-forward projection layer.

As in \cite{toutanova2015representing}, we use the best performing \textsc{DistMult} and \textsc{E+DistMult} as the base models. \textsc{DistMult}~\cite{yang2014embedding} learns latent vectors for the entities and each relation type, while model \textsc{E}~\cite{riedel2013relation} learns two latent vectors for each relation type, associated with its subject and object entities respectively. \textsc{E+DistMult} is a combination model that ensembles the predictions from individual models, and is trained jointly. We conduct experiments using only KB relations (\emph{KB only}), using their CNN to model textual relations (\emph{Conv}), and using our embedding to model textual relations (\emph{Emb}). 

The models are tested on predicting the object entities of a set of KB triples disjoint from the training set, given the subject entity and the relation type.
Table \ref{table:kbc_result} shows the performances of all models measured by mean reciprocal rank (MRR) of the correct entity, and HITS@10 (the percentage of test instances for which the correct entity is ranked within the top 10 predictions). We also show the performances on the two subsets of the test set, with and without textual mentions. The pre-trained embedding achieves comparable or better results to the CNN model trained with in-domain data. 

\footnotetext{The result of our implementation is slightly different from the original paper. We have communicated with the authors and agreed on the plausibility of the result. }

\begin{figure}[h]
\begin{center}
    \includegraphics[width=3.0in]{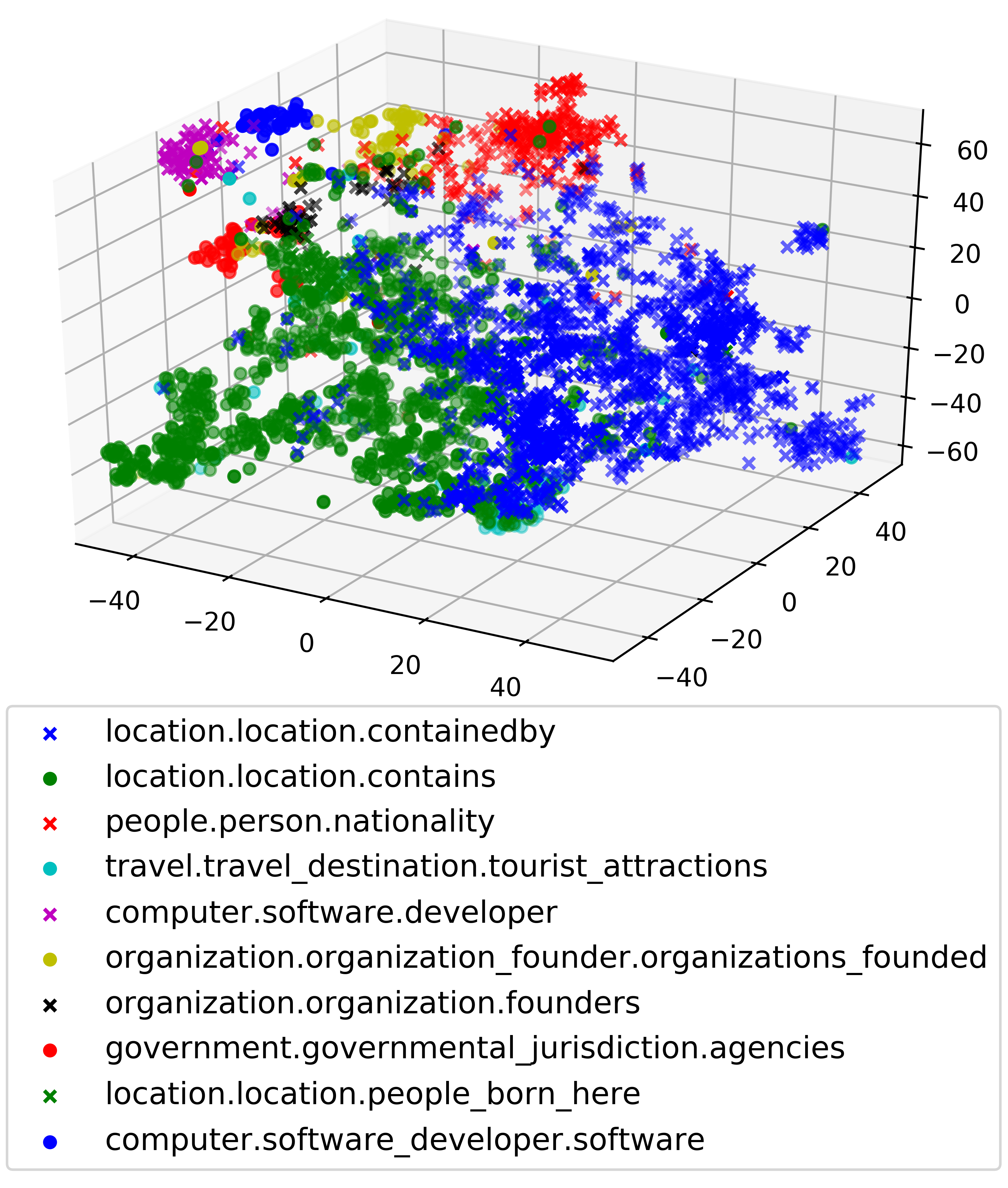}
    \caption{t-SNE visualization of our textual relation embeddings on ClueWeb validation data}
    \label{fig:tsne}
\end{center}
\end{figure}
\section{Analysis}
\label{sec:analysis}
\textbf{t-SNE visualization.} To measure the intrinsic property of the learned textual relation embedding, we apply t-SNE visualization \cite{maaten2008visualizing} on the learned embedding of ClueWeb validation set.

We filter out infrequent textual relations and assign labels to the textual relations when they co-occur more than half of the times with a KB relation. The visualization result of GloRE++ embedding associating with the top-10 frequent KB relations is shown in Figure \ref{fig:tsne}.
As we can see, similar textual relations are grouped together while dissimilar ones are separated. This implies that the embedding model can well generate textual relation representation for unseen textual relations, and can potentially serve as relational features to help tasks in unsupervised setting.

\textbf{Case Study.} To show that the embedding model generalizes to unseen textual relations via capturing crucial textual sub-patterns, we randomly pick some textual relations in NYT train set but not in ClueWeb train set, and compare with its top-5 nearest neighbors in ClueWeb train set, based on the similarity of the learned embedding. A case study is shown in Table \ref{table:case_2}.
We can see that the KB relation \textit{place\_of\_birth} often collocates with a preposition \textit{in} indicating the object fits into a location type, and some key words like \textit{born}. Together, the sub-structure \textit{born in} serves as a strong indicator for \textit{place\_of\_birth} relation.
There is almost always some redundant information in the textual relations, for example in the textual relation $\xleftarrow{nsubjpass} born \xrightarrow{nmod:on} nov. \xrightarrow{nmod:in}$, the sub-structure $\xrightarrow{nmod:on} nov.$ does not carry crucial information indicating the target relation. A good textual relation embedding model should be capable of learning to attend to the crucial semantic patterns. 

\section*{Acknowledgment}
The authors would like to thank the anonymous reviewers for their thoughtful comments. This research was sponsored in part by the Army Research Laboratory under cooperative agreements W911NF09-2-0053 and NSF IIS 1528175. The views and conclusions contained herein are those of the authors and should not be interpreted as representing the official policies, either expressed or implied, of the Army Research Laboratory or the U.S. Government. The U.S. Government is authorized to reproduce and distribute reprints for Government purposes notwithstanding any copyright notice herein.

\bibliography{acl2019}
\bibliographystyle{acl_natbib}

\end{document}